\documentclass[letterpaper]{article} % DO NOT CHANGE THIS
\usepackage{aaai25}  % DO NOT CHANGE THIS
\usepackage{times}  % DO NOT CHANGE THIS
\usepackage{helvet}  % DO NOT CHANGE THIS
\usepackage{courier}  % DO NOT CHANGE THIS
\usepackage[hyphens]{url}  % DO NOT CHANGE THIS
\usepackage{graphicx} % DO NOT CHANGE THIS
\usepackage{fontawesome5}
\usepackage{tikz}
\usepackage{algorithm}
\usepackage{algorithmic}
\usepackage{multirow}
\usepackage{booktabs}
\usepackage{makecell}
\usepackage{pifont}
\usepackage{amsmath}
\usepackage{amssymb} 
\usepackage{colortbl} 
\definecolor{mygray}{gray}{.85}
\newcommand{\cmark}{\ding{51}}%
\newcommand{\xmark}{\ding{55}}%
\usepackage[misc]{ifsym}

\usepackage{natbib}  % DO NOT CHANGE THIS AND DO NOT ADD ANY OPTIONS TO IT
\usepackage{caption} % DO NOT CHANGE THIS AND DO NOT ADD ANY OPTIONS TO IT
\usepackage{algorithm}
\usepackage{algorithmic}

\usepackage{newfloat}
\usepackage{listings}
\DeclareCaptionStyle{ruled}{labelfont=normalfont,labelsep=colon,strut=off} % DO NOT CHANGE THIS
\floatstyle{ruled}
\newfloat{listing}{tb}{lst}{}
\floatname{listing}{Listing}
\title{M3Net: Multimodal Multi-task Learning for 3D Detection, Segmentation, and Occupancy Prediction in Autonomous Driving}
\author{
    Xuesong Chen\textsuperscript{\rm 1},    
    Shaoshuai Shi\textsuperscript{\rm 1}\thanks{Corresponding author}\,
    Tao Ma\textsuperscript{\rm 1}, \\
    Jingqiu Zhou\textsuperscript{\rm 1},
    Simon See\textsuperscript{\rm 2}, 
    Ka Chun Cheung\textsuperscript{\rm 2},
    Hongsheng Li\textsuperscript{\rm 1, 3}
}
\affiliations{
    \textsuperscript{\rm 1} MMLab, The Chinese University of HongKong \\
    \textsuperscript{\rm 2} NVIDIA AI Technology Center \\
    \textsuperscript{\rm 3} Centre for Perceptual and Interactive Intelligence \\
    \{chenxuesong@link, hsli@ee\}.cuhk.edu.hk, ~shaoshuaics@gmail.com 

}

\usepackage{bibentry}
\begin{document}

\maketitle

\begin{abstract}
The perception system for autonomous driving generally requires to handle multiple diverse sub-tasks. However, current algorithms typically tackle individual sub-tasks separately, which leads to low efficiency when aiming at obtaining full-perception results. Some multi-task learning methods try to unify multiple tasks with one model, but do not solve the conflicts in multi-task learning.
In this paper, we introduce M3Net, a novel multimodal and multi-task network that simultaneously tackles detection, segmentation, and 3D occupancy prediction for autonomous driving and achieves superior performance than single task model. M3Net takes multimodal data as input and multiple tasks via query-token interactions. To enhance the integration of multi-modal features for multi-task learning, we first propose the \textbf{M}odality-\textbf{A}daptive \textbf{F}eature \textbf{I}ntegration (MAFI) module, which enables single-modality features to predict channel-wise attention weights for their high-performing tasks, respectively. Based on integrated features, we then develop task-specific query initialization strategies to accommodate the needs of detection/segmentation and 3D occupancy prediction. Leveraging the properly initialized queries, a shared decoder transforms queries and BEV features layer-wise, facilitating multi-task learning. Furthermore, we propose a \textbf{T}ask-oriented \textbf{C}hannel \textbf{S}caling (TCS) module in the decoder to mitigate conflicts between optimizing for different tasks. Additionally, our proposed multi-task querying and TCS module support both Transformer-based decoder and Mamba-based decoder, demonstrating its flexibility to different architectures.
M3Net achieves state-of-the-art multi-task learning performance on the nuScenes benchmarks. 
% Code is available at {https://github.com/Cedarch/M3Net}.
\end{abstract}

% Uncomment the following to link to your code, datasets, an extended version or similar.
%
\begin{links}
    \link{Code}{https://github.com/Cedarch/M3Net}
    % \link{Datasets}{https://aaai.org/example/datasets}
    % \link{Extended version}{https://aaai.org/example/extended-version}
\end{links}

\section{Introduction}

\label{sec:intro}
The perception module is important for autonomous driving system, which aims to accurately perceive the environment around the self-driving vehicle. To ensure reliable driving, the perception task is divided into different sub-tasks, including detection, map segmentation, occupancy prediction, etc. However, current algorithms generally tackle individual sub-tasks and thus require deploying multiple independent models to obtain full-perception results. This practice poses notable challenges: 1) It overlooks task similarities, leading to redundant computations, and 2) it impedes the mutual enhancement between the multiple sub-tasks. Therefore, a unified model that can simultaneously tackle multiple perception sub-tasks holds significant promise.

However, integrating multiple tasks on a fully shared backbone often results in low performance. This is because the feature representations required by different tasks vary, leading to gradient conflicts when training the network.
Various approaches have been developed to tackle the conflicts in multi-task learning, including shared feature representations~\cite{he2017mask,borsa2016learning}, task-specific layers~\cite{wallingford2022task}, and attention mechanisms~\cite{goncalves2023mtlsegformer,liu2019end}, etc. Shared feature representations involve sharing a backbone to utilize shared knowledge. 
Some methods~\cite{li2024omg,liu2023petrv2} utilize the query as a unified representation across tasks to get multi-task results. 
Task-specific layers equipped with attention mechanisms facilitate task-specific learning and enable the model to handle task-relevant information. Although these strategies perform well in 2D perception tasks, challenges arise in the context of 3D tasks in autonomous driving.

Specifically, tasks of object detection and map segmentation can be effectively tackled using Bird's Eye View (BEV) features, whereas 3D occupancy estimation requires computational-intensive 3D features for achieving dense 3D prediction. Most previous methods use BEV features to unify the representation of multiple tasks. However, they have difficulties in effectively tackling occupancy prediction and semantic conflicts from the compression of the height dimension in the BEV space. For example, a vehicle may share features with both the trees above it and the road below it. Consequently, the network struggles to learn features to handle multiple classes simultaneously. In response to this challenge, Univision~\cite{hong2024univision} maintains both 3D and BEV features and successfully handles detection and occupancy prediction tasks simultaneously, but it comes at the cost of efficiency.
Other methods, such as MetaBEV~\cite{ge2023metabev} and BEVFusion~\cite{liu2023bevfusion}, try to handle multi-task learning in one model but overlook the important 3D occupancy prediction task. They only focus on BEV-related tasks, including detection and map segmentation, through task-specific modules such as Mixture of Experts (MoE) and unshared decoders. 
Despite the efforts, the performance of multi-task learning for autonomous driving perception still falls short of single-task learning.

In this paper, a unified \textbf{M}ulti-\textbf{M}odal and \textbf{M}ulti-task learning network, named {\it M3Net}, is proposed which can handle multiple perception tasks in autonomous driving, including 3D objection detection, BEV map segmentation, and 3D occupancy prediction, and yields superior performance to independently trained single-task models.

M3Net first leverages the well-established backbone of BEVFusion to encode BEV features from camera and LiDAR sensors. Based on the BEV feature, M3Net then unifies multiple tasks into the form of query-token interactions and proposes to utilize 3D and BEV queries in a shared BEV feature decoder for multi-task learning.
To accommodate the representation differences between the 3D occupancy prediction and BEV-based detection and segmentation, we propose a task-specific query initialization strategy for different tasks. 
For detection and segmentation tasks, we choose specific features from the BEV feature map as the initial queries and utilize their positions as the queries' position embedding. 
For occupancy prediction, each voxel in the 3D space can act as an occupancy query, with its corresponding 
%3D position embedding. 

Although utilizing the proposed query initialization strategy, a powerful BEV feature that fully exploits the advantages of different modalities is still crucial for multi-task learning. For example, LiDAR point clouds are more suitable for detection tasks due to their precise geometric structure, while images contain more dense semantic information and thus perform better on segmentation tasks.
To this end, we propose a Modality-Adaptive Feature Integration (MAFI) module. This module enhances the fusion of multi-modal features by allowing single-modality features to predict channel-wise gating weights for their proficient tasks.
Specifically, the MAFI module initially fuses two modalities' features using a simple convolution layer, similar to the operation in BEVFusion. Subsequently, LiDAR and image BEV features independently predict lidar-adaptive and camera-adaptive weights to perform transformations on the initial fused features, resulting in modality-adapted multi-modal features respectively. Finally, these adapted features are added to form the final integrated BEV features.
Furthermore, the integrated BEV features still face challenges of gradient conflicts between the tasks during optimization. 
To address this challenge, we propose an effective and efficient \textbf{T}ask-oriented \textbf{C}hannel \textbf{S}caling (TCS) mechanism, which dynamically predicts channel scaling weights of the shared BEV features for different tasks from the same input. Subsequently, these weights are multiplied with the shared BEV features to derive task-specific BEV features. Through the proposed TCS, the queries of different tasks selectively gather information from the shared BEV feature and effectively mitigate conflicts between multiple tasks.
Following several decoder layers, each task's queries are fed into different task heads. With our MAFI and TCS modules, M3Net better integrates multi-modal features and tackles multiple perception tasks in a lightweight manner.

Our proposed M3Net supports both transformer-based and mamba-based decoders, featuring an attention mechanism with linear complexity, which demonstrates M3Net's flexibility and the effectiveness of the mamba mechanism in autonomous driving perception tasks.

In a nutshell, our contributions are four-fold: 
  1) We introduce a unified, multi-modal, and multi-task learning network M3Net, which can simultaneously tackle multiple perception tasks for autonomous driving.
  2) We propose a simple and effective modality-adaptive multi-modal BEV feature integration module to effectively leverage the strengths of different modality features.
  3) We propose the task-oriented channel scaling mechanism to alleviate gradient conflicts between different tasks during optimization, allowing M3Net to obtain higher performances than single-task models.
  4) Our framework is compatible with both transformer and mamba decoders, demonstrating its flexibility and mamba decoder's capability in 3D perception tasks.

\begin{figure*}[!t]
\centering
\includegraphics[width=0.95\textwidth ]{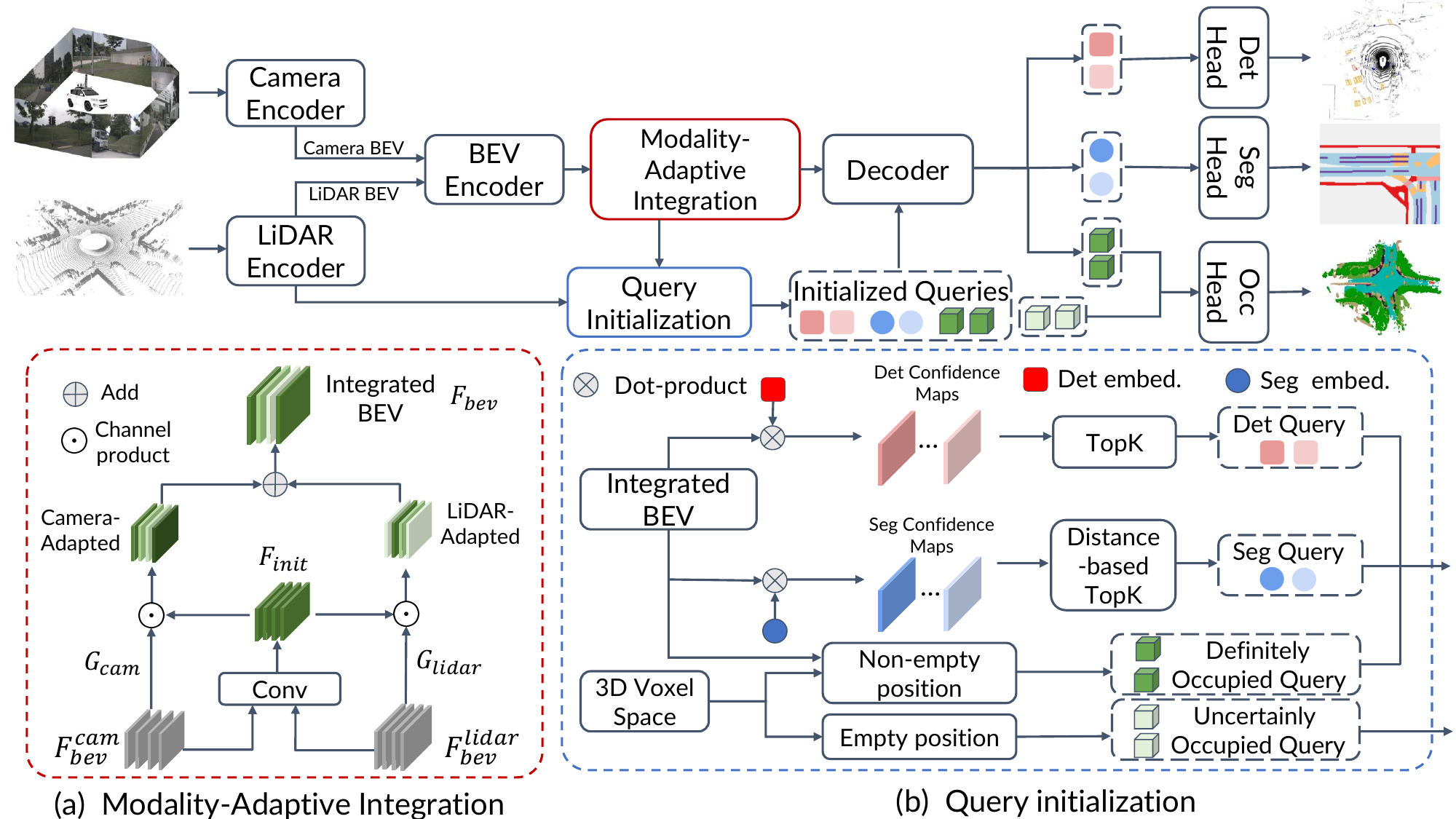}
\vspace{-2mm}
\caption{The overall architecture of our proposed M3Net as well as the detailed design of (a) Modality-adaptive Feature Integration module and (b) BEV-based multi-task query initialization.}
\label{fig:pipeline}
\vspace{-6mm}
\end{figure*}

\section{Related Work}
\noindent
\textbf{3D Object Detection and Semantic Map Segmentation.}
3D detection involves locating and classifying objects in the 3D world, while map segmentation entails classifying regions within a map, such as drive areas or pedestrian crossing lines. Methods in this field can be categorized into single-modal and multi-modal approaches based on their input data. For single-modal methods, BEVDet~\cite{huang2021bevdet}, BEVFormer~\cite{li2022bevformer} and BEVsegformer~\cite{peng2023bevsegformer} transform images from surround-view cameras into a single BEV feature map for 3D detection and segmentation. LiDAR-based methods~\cite{shi2020pv, chen2022mppnet,chen2023trajectoryformer} mainly focus on 3D detection from single-frame or multi-frame point clouds.
Multi-modal approaches~\cite{liu2023bevfusion, ge2023metabev} utilize inputs from multiple sensors and fuse them in BEV space, adopting strategies similar to LiDAR-based methods for 3D detection and map segmentation.

\noindent
\textbf{3D Occupancy Prediction.}
Occupancy prediction involves making judgments about whether a position in a 3D scene is occupied and identifying the occupation category. Early methods in this domain, such as~\cite{ song2017semanticssc, cai2021semantic}, relied on vision-based approaches utilizing images and additional geometric information to perform occupancy prediction. 
Recently, TPVFormer~\cite{huang2023tri} improves the representation by introducing the Tri-Perspective-View (TPV), which enriches the BEV with Z-axis information for a more detailed scene understanding.  OccFormer~\cite{zhang2023occformer} proposes a dual-path transformer network tailored to processing 3D volumes, specifically targeting semantic occupancy prediction tasks. VoxFormer~\cite{li2023voxformer} proposes a transformer-based two-stage network, where the first stage focuses on a sparse set of visible and occupied voxels, and the second stage generates dense 3D voxels from the sparse ones.
Additionally, OpenOccupancy~\cite{wang2023openoccupancy}, Occ3D~\cite{tian2024occ3d}, and SurroundOcc~\cite{wei2023surroundocc} have also contributed by devising pipelines dedicated to generating high-fidelity dense occupancy labels.

\noindent
\textbf{Multi-Task Learning.}
Multi-task frameworks aim to efficiently manage various tasks within a singular network. These methods are primarily categorized into two groups~\cite{huang2023fuller}: optimization techniques~\cite{chen2018gradnorm,mordan2021detecting,zhou2019finite, wang2019distributed} and network architecture enhancement\cite{ruder2019latent,guo2020learning}.
FULLER~\cite{huang2023fuller} proposes a multi-task optimization method to balance the loss weights of different tasks during training, preventing one task from dominating another. 
On the other hand, LidarMTL~\cite{feng2021simple} and LidarMultiNet~\cite{ye2023lidarmultinet} utilize a shared network for tasks encompassing 3D detection, segmentation, and road understanding based on point cloud data.
Meanwhile, \cite{wang2024panoocc} focuses on vision-centric (only using camera feature) multiple perception tasks in autonomous driving. 
In this paper, M3Net aims to use multimodal data within a unified architecture to address multiple tasks in autonomous driving.

\section{Methodology}

\subsection{Overview}
\label{sec:overview}
M3Net takes LiDAR point cloud and surrounding-view camera images as input and tackles the perception tasks of 3D object detection, BEV map segmentation, and 3D occupancy prediction simultaneously. To tackle multi-task learning, M3Net unifies the multiple perception tasks into the form of query-token interactions and adopts task-specific queries through a shared BEV feature decoder so that different perception tasks can be unified in one model. 

As illustrated in Fig.~\ref{fig:pipeline}, M3Net comprises four sequential sub-networks: a multimodal BEV feature encoder, a modality-adaptive feature integration module,
a multi-task query initialization network, and a multi-task decoder.

\subsection{Modality-adaptive BEV Feature Integration}
We unify both detection, segmentation, and 3D occupancy prediction into the BEV feature space, considering its efficiency, which is different from Univision~\cite{hong2024univision} utilizing both BEV and 3D features for multiple tasks.
To obtain BEV features,  we utilize the feature backbone of BEVFusion to process inputs from LiDAR and cameras and get corresponding BEV features, i.e., $F^{lidar}_{bev}$ and $F^{cam}_{bev}$. 
Subsequently, we introduce the Modality-adaptive Feature Integration (MAFI) module to optimally leverage the strengths of diverse modalities. As illustrated in Fig.~\ref{fig:pipeline} (a), the integration module begins with an initial fusion of $F^{lidar}_{bev}$ and $F^{cam}_{bev}$ features using a convolution layer, getting $F_{init}$.
Then, we employ an adaptive gating mechanism to integrate the advantages of LiDAR features in the detection task and the efficacy of image features in the segmentation task. Specifically, $F^{lidar}_{bev}$ and $F^{cam}_{bev}$ BEV features predict LiDAR-adaptive gating weights $G_{lidar}$ and camera-adaptive gating weights $G_{cam}$ through linear layers followed by a sigmoid activation function, respectively.

These weights enable distinct gating transformations on the initially fused features, resulting in adapted features. The final modality-adapted BEV features are formulated as:
\begin{equation}
\begin{aligned}
F_{bev} &= G_{lidar} \odot F_{init} + G_{cam} \odot F_{init}. \\
\end{aligned}
\end{equation}
 
Because of the better utilization of the advantages of single-modal features, the dense 3D Occupancy prediction task, which combines the information of both foreground detection and background segmentation, can also benefit from our MAFI module. 
With $F_{bev}$ in hand, the multi-task queries can be better initialized from the integrated feature.

\subsection{BEV-based Multi-task Query Initialization}
\label{sec:query_init}
Query initialization is vital for query-based networks to achieve satisfactory performance. Our experiments show that randomly initialized queries do not carry explicit semantic features and lead to inferior performance. To this end, we propose query initialization strategies based on multimodal BEV features to take advantage of the semantic and location information encoded in the BEV feature map. 

\noindent
\textbf{Occupancy query initialization.} 
We generate a 3D voxel feature space of dimensions $(H\times W \times Z)$. Each voxel in this space serves as a query, amounting to a total of $H\times W\times Z$ $C$-dimensional occupancy queries. We employ a Multi-Layer Perceptron (MLP) to encode 3D positions into their corresponding $C$-dimensional position embeddings. Given that most 3D occupancy queries are unoccupied, we utilize LiDAR point location information to categorize 3D occupancy queries into two groups: definitely occupied (with LiDAR points in the voxel) and uncertainly occupied (without LiDAR points in the voxel). Then different initialization strategies are adopted for them.
Specifically, we project the 2D BEV feature to their corresponding 3D positions as the initial feature of the definitely occupied queries.
Note that queries at the same 2D position but of varying heights share the same BEV-feature queries but with different positional embeddings.
On the other hand, for uncertainly occupied queries, a shared and learnable embedding is employed as their initial feature. 
At last, we add the queries' initial features and positional embedding to obtain the final initialized occupancy queries.

\noindent
\textbf{Detection and segmentation query initialization.} 
We choose corresponding features from the integrated multimodal BEV feature map, which are added with positional embeddings, as the initial queries for the detection and segmentation tasks. Such queries are naturally better aligned with the BEV feature map and would be later used in the decoder's cross-attention blocks to perform task-specific transformations. 
To select better locations to generate queries from the BEV feature map, we employ a unified segmentation approach to predict confidence maps for both object centers and map segmentation.
Specifically, we first randomly initialize 16 category embeddings, including 10 detection categories and 6 segmentation categories, and then construct a detection embedding map and a segmentation embedding map.
Subsequently, following Mask2Former~\cite{cheng2021mask2former}, we apply the dot-product between each category embedding and the corresponding embedding map to obtain the confidence maps.
With these confidence maps, for the detection task, we employ $n_d$ BEV features by selecting top-${n_d}$ candidates from the $H\times W\times 10$ confidence space as detection queries.
For the segmentation task, considering the distribution of LiDAR points varies significantly along the forward direction of the vehicle, we introduce a distance-based segmentation query initialization strategy. Specifically, we partition the confidence map evenly into $S$ blocks (corresponding to different distance ranges) along the forward direction of the vehicle. Within each block, for each segmentation class, the BEV feature with the highest confidence is selected as the query, resulting in a total of $n_s=S\times 6$ segmentation queries. 
To this end, queries of different blocks are tasked with segmenting the map within its corresponding distance range. 

\begin{figure}[!t]
\centering
\includegraphics[width=0.45\textwidth ]{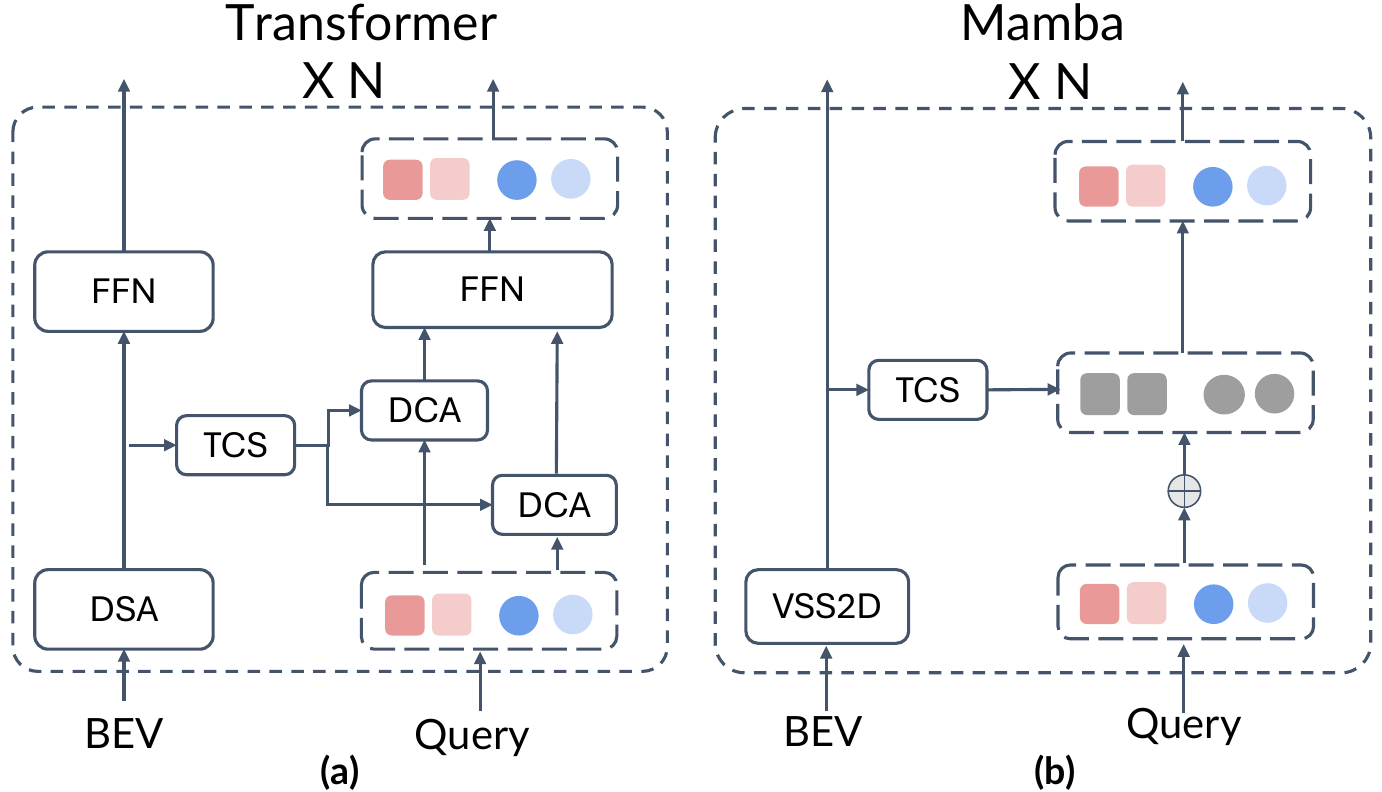}
\vspace{-2mm}
\caption{The detailed architectures of our transformer-based and mamba-based decoder layer with the task-oriented channel scaling module. DSA, DCA and VSS2D denote deformable self-attention, cross-attention and the VSS2D block from Vmamba~\cite{liu2024vmamba}.} 
\label{fig:decoder}
\vspace{-7mm}
\end{figure}

% \vspace{-3mm}
\subsection{Decoder with Task-oriented Channel Scaling}
\vspace{-1mm}
\label{sec:decoder}
Following query initialization, the decoder iteratively updates the multi-task queries and BEV features in a layer-by-layer manner. Notably, only the definitely occupied occupancy queries, detection queries, and segmentation queries are input into the decoder. The uncertainly occupied occupancy queries subsequently interact with the definitely occupied occupancy queries via attention in the occupancy prediction head, jointly predicting the dense 3D occupancy volume. Our decoder supports both transformer-based and mamba-based attention mechanisms for multi-task learning.

\noindent
\textbf{Transformer-based Decoder.} As shown in Fig. \ref{fig:decoder} (a), in each layer of the transformer-based decoder, we utilize the deformable self-attention layer \cite{zhu2020deformable} to propagate information between all features in the BEV space, so that each BEV feature can obtain long-range semantic relations in the whole context. 
After the updating of the BEV features, task-specific queries can aggregate richer information from the BEV space through a cross-attention layer. However, conflicts are inevitable when queries attempt to aggregate task-specific information from the shared BEV. To this end, we introduce a Task-specific Channel Scaling (TCS) operation to obtain task-specific BEV features. 

\noindent
\textbf{Task-oriented Channel Scaling.} 
We propose the Task-oriented Channel Scaling (TCS) mechanism to perform channel scaling of the shared BEV features so that task-specific queries can selectively gather different information to alleviate conflicts between multiple tasks. 
Specifically, taking the BEV feature $F_{bev}$ derived from an FFN module as input, we employ $n$ separate Linear layers followed by an activation function to obtain $n$ task-specific embeddings, where $n$ is the number of tasks and is 3 in our experiments. Then, these embeddings are forwarded into $n$ different linear layers to obtain $n$ task-specific channel scaling weights $W^{i} \in \mathbb{R}^{H\times W\times C}$, where ${i = 1,\dots,n}$. Multiplying the shared BEV features by these scaling weights: %yields task-specific BEV features:

\begin{equation}
F^{i}_{bev}=W^{i} \cdot F_{bev},  \text{~for~} i = 1, \dots, n.
\label{eq:tcs}
\end{equation}
Subsequently, task-specific queries can perform cross-attention from corresponding tasks' $F^{i}_{bev}$.

\noindent
\textbf{Mamba-based Decoder.} The Mamba~\cite{gu2023mamba} network introduces State Space Models (SSMs) to manage attention computations with linear complexity, capturing long-range dependencies while preserving computational efficiency. 
We adopt the design of Vmamba~\cite{liu2024vmamba} to scan each BEV feature from different directions. For each BEV location, we scan information from its surrounding positions in four different directions: top, bottom, left, and right. Subsequently, we merge the features scanned from these directions to update the current position's feature.
Utilizing the scan and merge operation, the Vision Selective Scan 2D (VSS2D) block, proposed in Vmamba,
can perform spatial information propagation in BEV features, $F_{mbev} = \text{VSS2D}(F_{bev})$.
Then, similar to the transformer-based decoder, we use TCS operation to obtain the task-specific BEV feature $F^i_{mbev}$.
Then multi-task queries can interact with $F^i_{mbev}$ to update their features, as shown in Fig.~\ref{fig:decoder} (b). Specifically, we adopt the index operation to obtain BEV features from the corresponding positions of different queries to update their feature, expressed as $q^i = q^i + \text{Index} (F^i_{mbev})$,
where $q^{i}$ means task-specific queries and $i\in \{1, \cdots, n\}$.
After updating queries and BEV features within one layer, we stack several layers to gradually complete multi-task learning for both transformer-based and mamba-based decoders.

\subsection{Multi-task Heads and Losses}

For the detection task, we use the center-based head~\cite{bai2022transfusion} and the detection loss is composed of the classification loss $L_{cls}$ and the regression loss $L_{reg}$,denoted as $L_{\text {det}}=\lambda_1 \cdot L_{cls}+\lambda_2 \cdot L_{reg}$, where $\lambda_1, \lambda_{2}$ denote hyper-parameters.
For the segmentation task, the same operation for the map segmentation confidence prediction is adopted.  We utilize the segmentation-specific BEV feature $F^i_{bev}$ from the last decoder layer as the embedding and we dot-product this embedding with segmentation queries, obtaining map segmentation results. Following BEVFusion, we use the standard focal loss~\cite{lin2017focal} as segmentation loss, denoted as $L{seg}$. 
For the 3D occupancy prediction task, we first merge the updated definitely occupied queries with uncertainly occupied queries to derive the initial voxel features $F_{3D} \in \mathbb{R}^{H\times W 
\times Z\times C}$ to obtain dense 3D voxel features. Subsequently, we employ a transformer or mamba block to refine the voxel features. 
After obtaining refined voxel features, we upsample it and project it to the output space to get the final results (the same shape with ground truth) $O \in \mathbb{R}^{H^{'}\times W^{'}\times Z^{'}\times (M+1)}$, where $M+1$ denotes $M$ semantic classes and one empty class. 
We follow the loss setting in OpenOccupancy~\cite{wang2023openoccupancy}, denoted as $\mathcal{L}_{\text {occ}}$.
The total loss is formulated as:
$L = \lambda_{det} \cdot L_{\text {det}} +\lambda_{seg} \cdot L_{\text {seg}} + \lambda_{occ} \cdot L_{\text {occ}}$, 
where $\lambda_{det},\lambda_{seg},\lambda_{occ}$ denote hyper-parameters.

\section{Experiments}

\subsection{Dataset and Implementation Details}
\noindent
\textbf{Dataset.} 
For detection and segmentation, we evaluate M3Net with the nuScenes~\cite{caesar2020nuscenes} dataset, a large-scale multimodal dataset designed for 3D detection and map segmentation. 
It encompasses data from various sensors, including 6 cameras, one LiDAR, and five radars. 
We resize the input views to $256\times704$ resolution and voxelize the point cloud to 0.075m for detection, occupancy prediction, and segmentation tasks. For 3D detection evaluation, we employ the standard nuScenes Detection Score (NDS) and mean Average Precision (mAP). Additionally, for BEV map segmentation evaluation, we utilize the Mean Intersection over Union (mIoU) metric across all six categories, following the setting of BEVFusion~\cite{liu2023bevfusion}.
For occupancy prediction, we evaluate M3Net on OpenOccupancy~\cite{wang2023openoccupancy} dataset.
The dataset conducts occupancy annotation on the observed scene using multimodal input from the nuScenes dataset.

\noindent
\textbf{Implementation Details.} 
The implementation of M3Net is built upon the OpenPCDet framework~\cite{od2020openpcdet}.
For optimization, M3Net employs the AdamW~\cite{loshchilov2017decoupled} optimizer and cycle learning rate (lr) schedule with max lr setting to 0.0008. To save training resources, we initially train the detection and segmentation tasks for 12 epochs with CBGS sampling~\cite{zhu2019class} and then integrate the occupancy prediction task for the next 15 epochs' joint training without CBGS sampling. For hyper-parameters, We use $n_d=200$ detection queries, $n_s=30$ segmentation queries, and  $180\times 180\times 5$ occupancy prediction queries. The channel of all queries is 256. We set the number of decoder layers to 6 and the shape of the BEV feature is $180\times 180\times 256$. The proposed transformer-based and mamba-based M3Net share the same training setting.

\begin{table*}[!t]
\begin{center}
\resizebox{\linewidth}{!}{
    \begin{tabular}{lc | c| cc |ccccccc}
        \toprule
        Methods &  Modality & MTL & mAP(val) & NDS(val) & Drivable & Ped.Cross & Walkway & Stop Line & Carpark & Divider & Mean\\
        \midrule
        M\({}^{\mbox{2}}\)BEV\cite{Xie-22arxiv-m2bev}   & C & \xmark  & 41.7 & 47.0     & 77.2 & -    & -    & -    & -    & 40.5 & - \\
        BEVFormer\cite{li-22eccv-bevformer}             & C & \xmark & 41.6 & 51.7      & 80.1 & - & - & - & - & 25.7 & - \\

        BEVFusion\cite{liu2023bevfusion}              & C & \xmark & 35.6 & 41.2      & 81.7 & 54.8 & 58.4 & 47.4 & 50.7 & 46.4 & 56.6\\
        X-Align\cite{Borse-23wacv-xalign}               & C & \xmark & - & -            & 82.4 & 55.6 & 59.3 & 49.6 & 53.8 & 47.4 & 58.0 \\

        \midrule
        PointPillars\cite{lang-22cvpr-pointpillar}      & L & \xmark & 52.3 & 61.3       & 72.0 & 43.1 & 53.1 & 29.7 & 27.7 & 37.5 & 43.8\\
        CenterPoint\cite{yin-21cvpr-centerpoint}        & L & \xmark & 59.6 & 66.8       & 75.6 & 48.4 & 57.5 & 36.5 & 31.7 & 41.9 & 48.6\\
        BEVFusion\cite{liu2023bevfusion}              & L & \xmark & 64.7 & 69.3       & 75.6 & 48.4 & 57.5 & 36.4 & 31.7 & 41.9 & 48.6\\

        \midrule
        PointPainting\cite{vora-20cvpr-pointpainting}   & L+C & \xmark & 65.8 & 69.6     & 75.9 & 48.5 & 57.1 & 36.9 & 34.5 & 41.9 & 49.1\\
        MVP\cite{yin-21nips-mvp}                        & L+C & \xmark & 66.1 & 70.0     & 76.1 & 48.7 & 57.0 & 36.9 & 33.0 & 42.2 & 49.0\\
        TransFusion\cite{bai-22cvpr-transfusion}        & L+C & \xmark & 67.3 & 71.2     & - & - & - & - & - & - & - \\
        FULLER~\cite{huang2023fuller} & L+C & \xmark & 67.6 & 71.3 & - & - & - & - & - & - & 62.3 \\
        BEVFusion\cite{liu2023bevfusion}              & L+C & \xmark & 68.5 & 71.4     & 85.5 & 60.5 & 67.6 & 52.0 & 57.0 & 53.7 & 62.7\\
        X-Align\cite{Borse-23wacv-xalign}              & L+C & \xmark & - & - & 86.8 & 65.2 & 70.0 & 58.3 & 57.1 & 58.2 & 65.7 \\

        \midrule
        FULLER~\cite{huang2023fuller} & L+C & \cmark & 60.5 & 65.3 & - & - & - & - & - & - & 58.4 \\
        BEVFusion-MTL\cite{liu2023bevfusion} (share)         & L+C & \cmark & - & 69.7 & - & - & - & - & - & - & 54.0 \\
        BEVFusion-MTL\cite{liu2023bevfusion} (sep)         & L+C & \cmark & 65.8 & 69.8 & 83.9 & 55.7 & 63.8 & 43.4 & 54.8 & 49.6 & 58.5\\

        MetaBEV~\cite{ge2023metabev}(share)    & L+C & \cmark & 65.6 & 69.5 & 88.7 & 64.8 & 71.5 & 56.1 & 58.7 & 58.1 & 66.3 \\

        MetaBEV~\cite{ge2023metabev} (sep)   & L+C & \cmark & 65.4 & 69.8 & 88.5 & 64.9 & 71.8 & 56.7 & 61.1 & 58.2 & 66.9 \\
         \rowcolor{gray!25}
        M3Net (mamba)      & L+C & \cmark & 68.5 & 71.8 & \textbf{90.6} & 69.1 & \textbf{75.8} & 62.5 & \textbf{65.3} & \textbf{61.7} & \textbf{70.8} \\
        \rowcolor{gray!25}
        M3Net (transformer)      & L+C & \cmark & \textbf{69.0} &  \textbf{72.4} & 90.3 & \textbf{69.6} & 75.8 & \textbf{63.4} & 62.3 & 61.1 & 70.4 \\
        \bottomrule
    \end{tabular}
}
\vspace{-2mm}
\caption{\textbf{Performance comparisons on nuScenes val set.} ``share'' means multi-task heads share one BEV encoder and ``sep'' means task heads have separate encoders. M3Net is built upon the backbone of of BEVFusion.}
\label{tab:nuscenes_det_val}
\vspace{-0.2cm}
\end{center}
\end{table*} 

\begin{table*}[!t]
\setlength{\tabcolsep}{0.0035\linewidth}
\newcommand{\classfreq}[1]{{~\tiny(\semkitfreq{#1}\%)}}  %
\centering
\vspace{-2mm}
\resizebox{1\linewidth}{!}{
	\begin{tabular}{l|c c c | c | c c c c c c c c c c c c c c c c}
 
		\toprule
		Method
        & \makecell[c]{MTL}
		& \makecell[c]{Input}
		& \makecell[c]{Surround}
		& \makecell[c]{\colorbox[gray]{0.85}{mIoU}}
		& \rotatebox{90}{barrier} 
		& \rotatebox{90}{bicycle}
		& \rotatebox{90}{bus} 
		& \rotatebox{90}{car} 
		& \rotatebox{90}{const. veh.} 
		& \rotatebox{90}{motorcycle} 
		& \rotatebox{90}{pedestrian} 
		& \rotatebox{90}{traffic cone} 
		& \rotatebox{90}{trailer} 
		& \rotatebox{90}{truck} 
		& \rotatebox{90}{drive. suf.} 
		& \rotatebox{90}{other flat} 
		& \rotatebox{90}{sidewalk} 
		& \rotatebox{90}{terrain} 
		& \rotatebox{90}{manmade} 
		& \rotatebox{90}{vegetation} \\
		\midrule
		MonoScene~\cite{monoscene} & \xmark & C & \xmark  &  \cellcolor{mygray}6.9 & 7.1  & 3.9  &  9.3 &  7.2 & 5.6  & 3.0  &  5.9& 4.4& 4.9 & 4.2 & 14.9 & 6.3  & 7.9 & 7.4  & 10.0 & 7.6 \\
  
  		TPVFormer~\cite{tpv} & \xmark  &C &  \cmark& \cellcolor{mygray}7.8 & 9.3  & 4.1  &  11.3 &  10.1 & 5.2  & 4.3  & 5.9 & 5.3&  6.8& 6.5 & 13.6 & 9.0  & 8.3 & 8.0  & 9.2 & 8.2 \\

            AICNet~\cite{aicnet} & \xmark & C\&D   &  \xmark& \cellcolor{mygray}10.6 & 11.5  & 4.0  & 11.8  & 12.3&  5.1 & 3.8  & 6.2  & 6.0 & 8.2&  7.5&  24.1 & 13.0 & 12.8  & 11.5 & 11.6  &  20.2\\

            3DSketch~\cite{sketch} & \xmark &  C\&D & \xmark& \cellcolor{mygray}10.7  & 12.0 & 5.1&  10.7&  12.4& 6.5 & 4.0 & 5.0 & 6.3 & 8.0 & 7.2 &21.8 &14.8 &13.0 &11.8 &12.0 &21.2 \\
        
            LMSCNet~\cite{lmscnet} & \xmark & L &  \cmark& \cellcolor{mygray}11.5 & 12.4&  4.2 & 12.8  & 12.1  & 6.2  &  4.7 & 6.2 & 6.3&  8.8&  7.2& 24.2 & 12.3  & 16.6 & 14.1  & 13.9 & 22.2 \\

		JS3C-Net~\cite{js3cnet} & \xmark &L &  \cmark& \cellcolor{mygray}12.5 & 14.2 & 3.4  & 13.6  & 12.0  & 7.2  &  4.3 & 7.3 & 6.8&  9.2& 9.1 & 27.9 & 15.3  & 14.9 & 16.2  & 14.0 & 24.9 \\

            CONet~\cite{wang2023openoccupancy} (LR) & \xmark & C\&L &   \cmark& \cellcolor{mygray}15.1 &  14.3  & 12.0  & 15.2  & 14.9  & 13.7  & 15.0  & 13.1 & 9.0 & 10.0 & 14.5 & 23.2 & 17.5 & 16.1  & 17.2 & 15.3  & 19.5  \\

            CONet~\cite{wang2023openoccupancy} (HR) & \xmark & C\&L &   \cmark & \cellcolor{mygray}20.1 & 23.3  & 13.3  & 21.2  & 24.3  & 15.3  & 15.9  & 18.0 & 13.3 & 15.3 &20.7 & \textbf{33.2} & 21.0 & 22.5  & 21.5 & 19.6  & 23.2  \\
            \midrule
            \rowcolor{gray!25}
            M3Net (transformer) & \cmark & C\&L &   \cmark &\cellcolor{mygray} 23.3 &  27.5  & 19.3  & \textbf{23.6}  & 27.0  & \textbf{15.5}  & 28.5 & 34.1 & 19.1 & 17.7 & 23.0 & 27.8 & 16.1 & 22.0  & 21.0 & 24.1  & 26.5  \\
            \rowcolor{gray!25}
            M3Net (mamba)& \cmark & C\&L &   \cmark& \cellcolor{mygray} \textbf{24.1} & \textbf{28.7} & \textbf{20.1}  & 23.1  & \textbf{27.3}  & 15.1 & \textbf{28.6} & \textbf{34.6}  & \textbf{20.5} & \textbf{18.0} & \textbf{23.0} & 28.3 & \textbf{21.7} & \textbf{22.5} & \textbf{22.0}  & \textbf{24.7} & \textbf{27.2}   \\

		\bottomrule
	\end{tabular}
     }
\vspace{-2mm}
\caption{\textbf{Performance comparison on nuScenes-Occupancy val set.} The $C,D,L$ denotes camera, depth, LiDAR and $MTL$ means Multi-Task Learning. $LR$ means low resolution feature and $HR$ denotes high resolution feature.}
\label{tab:occ}
\vspace{-6mm}
\center
\end{table*}

\vspace{-3mm}
\subsection{Performance for Detection and Segmentation}
\vspace{-1mm}
Table~\ref{tab:nuscenes_det_val} shows the detection and segmentation results of state-of-the-art methods on nuscenes datasets. We present the performances of M3Net based on both transformer and mamba decoders.
For transformer-based M3Net, it shows consistency performance enhancements across detection and segmentation tasks, compared to independently trained BEVFusion~\cite{liu2023bevfusion}, achieving improvements of 0.5\%, 1.0\%, and 7.7\% in mAP, NDS, and IoU, respectively, which demonstrates that M3Net better utilizes the information of BEV features. Meanwhile, compared with BEVFusion-MTL ~\cite{liu2023bevfusion}, with two separate BEV encoders, named BEVFusion-MTL (sep), M3Net significantly exceed it by 3.2\%, 2.6\%, and 11.9\% in mAP, NDS, and IoU, proving the effectiveness of our multi-task learning strategy.
Moreover, the substantial enhancement in segmentation tasks further demonstrates the effectiveness of our strategy in unifying detection and segmentation tasks into query representation, contrasting with BEVFusion's strategy that adopts separate query-based detection and convolution-based segmentation heads.
When compared to the state-of-the-art multi-task learning method MetaBEV~\cite{ge2023metabev}, which also uses the backbone of BEVFusion, our approach surpasses its shared encoder version by 3.4\%, 2.9\%, and 3.1\% in mAP, NDS, and IoU metrics, respectively. Furthermore, even when compared to the separate encoder version, our method still outperforms it by 3.6\%, 2.6\%, and 3.5\% in mAP, NDS and IoU metrics. 
These results demonstrate the effectiveness of our MAFI module in better integrating multimodal features and our TCS operation in mitigating conflicts of multi-task learning, outperforming the MoE strategy utilized by MetaBEV. 

For mamba-based M3Net, it shows similar performance to our transformer model and surpasses the MetaBEV (sep), by 3.1\%, 2.0\%, and 3.9\% in terms of mAP, NDS, and IoU, demonstrating the flexibility of M3Net to different attention mechanisms. Meanwhile, the mamba model has fewer parameters than the transformer-based one by removing FFN.

\vspace{-0.3cm}
\subsection{Performance for 3D Occupancy Prediction}

Table~\ref{tab:occ} shows the results of 3D occupancy prediction of different methods on the nuScenes-based OpenOccupany~\cite{wang2023openoccupancy} dataset. Our transformer-based and mamba-based M3Net significantly surpass the state-of-the-art method CONet (HR) by 3.2\% and 4.0\% respectively in terms of mIoU. 
Thanks to the joint training with the detection task, M3Net has achieved great improvements in occupancy prediction of foreground objects, especially small objects. For example, mamba-based M3Net significantly surpasses CONet (HR) by about 12\% and 16\% on motorcycle and pedestrian, respectively, proving the superiority of multi-task learning.
On the other hand, the resolution of our adopted 3D feature space is much lower than that of CONet (HR), which utilizes the resolution of $512\times 512\times 40$ (the same as ground truth's resolution). However, M3Net works on a 3D feature space of resolution $180\times 180\times 5$, and obtains the 
 $512\times 512\times 40$ prediction results via up-sampling, which greatly reduces the computational load.

\subsection{Ablation Studies}
We conduct ablation studies to verify the effectiveness of each component of M3Net. To save the training cost, we initially train M3Net for 6 epochs for detection and segmentation, and then add the occupancy prediction task to jointly fine-tune the three tasks for another 10 epochs. All experiment results are trained on the transformer-based M3Net.

\noindent
\textbf{Multi-task model vs. single-task models.}
As shown in the 3rd to 5th rows of Table~\ref{tab:abla_mtl}, we independently train the M3Net-based single-task models using the same training epochs of multi-task learning. Then we train M3Net without our proposed TCS module and MAFI module but with our proposed multi-task query initialization, named M3Net multi-query,  to perform multi-task learning, as shown in the 6th row in Table~\ref{tab:abla_mtl}.
Without TCS and MAFI modules, although different tasks are assigned to task-specific queries, conflicts between tasks still exist on the shared BEV features, resulting in the detection and segmentation tasks performance loss of 0.8\%, 0.6\% (total 1.4\%) in terms of NDS and IoU. However, this performance loss is much lower than that of BEVFusion, where NDS and IoU decrease by 0.7\% and 8.7\% (total 9.4\%) respectively. On the other hand, our M3Net multi-query baseline still achieves comparable detection performance and higher segmentation performance than independently trained BEVFusion, demonstrating the effectiveness of our proposed approach of tackling multiple tasks simultaneously with task-specific query-token interaction in a unified model.
For the occupancy, there is a notable improvement of 6\%  on mIoU to single-task model because the occupancy queries can leverage shared knowledge.

\begin{table}[!t]
\centering
\resizebox{0.99\linewidth}{!}{
\begin{tabular}{l|c|c|c|c}
\hline
\multirow{2}{*}{Method} & Det & Seg & Occ & \multicolumn{1}{c}{\multirow{2}{*}{$\triangle$ MTL}}
\\
& \footnotesize{NDS}    & \footnotesize{IoU}  & \footnotesize{mIoU} &     \\ \hline
BEVFusion detection-only        & 71.4          & *            & *         &*         \\
BEVFusion segmentation-only        &           & 62.7            & *         &*         \\ \hline \hline
M3Net detection-only              & 72.0          & *            & *         &*         \\
M3Net segmentation-only     & *             & 66.3         & *               &*          \\
M3Net occupancy-only        & *             & *            & 15.7            &*          \\ \hline \hline
M3Net multi-query           & 71.2          & 65.7         & 21.7          & +4.6                   \\
M3Net multi-query + MAFI        & 71.5          & 66.3         & 22.1        & +5.9        \\
\rowcolor{gray!25}
M3Net multi-query + MAFI + TCS        & 72.2          & 66.7         & 22.4        & +7.3                    \\ \hline
\end{tabular}
}
\vspace{-2mm}
\caption{Performances of M3Net based single-task models and different multi-task learning methods. $\triangle$ MTL represents the sum of the improvements of the proposed method over the independently trained single-task models. }
\label{tab:abla_mtl}
\vspace{-6mm}
\end{table}

\begin{table}[!t]
\begin{minipage}[b]{0.47\linewidth}
\centering
\vspace{-3mm}
\resizebox{0.99\textwidth}{!}{
\begin{tabular}{c|c|c}
\hline
Query No.  & Layout  & IoU \\ \hline
6   &  $1\times 6$       &  65.6   \\
\rowcolor{gray!25}
30  &  $5\times 6$      &  66.3  \\
60  &  $10\times 6$     &  66.4   \\ \hline
\end{tabular}}
\caption{Performance comparison of different segmentation query init. strategies. }
\label{tab:abla_seg_query}
\end{minipage}
\hspace{10pt}
\begin{minipage}[b]{0.47\linewidth}
\centering
\vspace{5mm}
\resizebox{1.0\textwidth}{!}{
\begin{tabular}{c|c|c}
\hline
Method   & FFN & Para.(\textbf{M}) \\ \hline
Tr. Dec & \checkmark   &   18.9         \\
Ma. Dec  & \xmark  &   9.0       \\ \hline
\end{tabular}
}
\vspace{3mm}
\caption{Comparison of parameters for different attention mechanisms in decoder.}
\label{tab:abla_cost}
\end{minipage}
\vspace{-9mm}
\end{table}

\noindent
\textbf{Effects of the modality-adaptive feature integration.}
The 7th row (M3Net multi-query + MAFI) of Table~\ref{tab:abla_mtl}, shows the effects of the MAFI module. The MAFI module brings improvements of 0.3\%, 0.6\%, and 0.4\% in terms of NDS, IoU, and mIoU, respectively, to the multi-query baseline and a total performance improvement $\triangle$ MTL of 5.9\% for single-task models. Allowing single-modality features to predict channel-wise gating weights for their proficient tasks, the MAFI module better preserves the features that are beneficial to different single tasks during feature integration. As a result, the 3D occupancy prediction task also benefits from better-integrated BEV features.

\noindent
\textbf{Effects of the task-oriented channel scaling.}
The results in Table~\ref{tab:abla_mtl} demonstrate the TCS module's efficacy in mitigating multi-task learning conflicts. Building upon MAFI's multimodal feature integration, TCS further elevates performance by 7.3\% ($\triangle$ MTL) compared to single-task models. Specifically, TCS improves NDS by 0.7\% in detection, IoU by 0.4\% in segmentation, and mIoU by 0.3\% in occupancy prediction relative to the MAFI baseline. Notably, these metrics surpass their single-task counterparts, with occupancy prediction showing a substantial 6.7\% mIoU gain. These improvements validate TCS's effectiveness.

\noindent
\textbf{Effects of the segmentation query initialization.}
Table~\ref{tab:abla_seg_query} evaluates our distance-based query initialization strategy for segmentation. The query layout is represented as (M, N), where M denotes the number of map blocks and N indicates the class count. The 6-query configuration assigns one query per category for global (100m×100m) segmentation. The 30-query setting divides the range into 5 blocks, with 6 queries per (20m×100m) block. Results show that the 6-query configuration underperforms due to limited global semantic representation capacity. The 30-query setting improves performance by 0.7\%, validating our distance-based initialization approach. The marginal 0.1\% gain from 60 queries leads us to adopt 30 queries as the optimal parameter-performance trade-off.

\noindent
\textbf{Effects of the segmentation query initialization.}
\textbf{Analysis of transformer-based and mamba-based decoders.}
We present the parameter amount of transformer-based (Tr. Dec) and mamba-based (Ma. Dec) decoder of M3Net in Table~\ref{tab:abla_cost}.
The mamba decoder exhibits significantly fewer parameters compared to the deformable transformer attention, primarily due to its independence from FFN. Meanwhile, our mamba-based M3Net achieves comparable performance to the transformer for detection task and obtains better results for map segmentation and occupancy prediction tasks, demonstrating the potential of using a mamba decoder in addressing perception tasks for autonomous driving.

\section{Conclusion}
In this work, we propose a multi-modal and multi-task network M3Net, which unifies different tasks into a query-token interaction manner and can tackle 3D objection, map segmentation, and 3D occupancy prediction simultaneously. We first propose a modality-adaptive feature integration module to better combine the advantages of single-modal features and obtain the modality-adapted multi-modal BEV feature. Based on the integrated BEV feature, we propose a BEV-based task-specific query initialization strategy to accommodate the representation disparities for different tasks. With the initialized queries, we adopt a shared decoder to update queries, facilitating multi-task learning. Furthermore, a task-oriented channel scaling module is proposed to tackle the conflicts among tasks. 
The experiments on nuScenes datasets demonstrate that our approach effectively integrates the features of the two modalities and alleviates the conflicts of different tasks, outperforming state-of-the-art multi-task learning methods with remarkable margins.

\section{Acknowledgements}
This project is funded in part by National Key R\&D Program of China Project 2022ZD0161100, by Shanghai Artificial Intelligence Laboratory (Grant No. 2022ZD0160104), by the Centre for Perceptual and Interactive Intelligence (CPII) Ltd under the Innovation and Technology Commission (ITC)’s InnoHK, by General Research Fund of Hong Kong RGC Project 14204021, by Smart Traffic Fund PSRI/76/2311/PR for algorithm framework design and dataset curation. Hongsheng Li is a PI of CPII under the InnoHK.

\bibliography{aaai25}

\end{document}